\PassOptionsToPackage{table}{xcolor}
\documentclass[sigconf]{acmart}

\AtBeginDocument{%
  }

\usepackage{times}
\usepackage{latexsym}
\usepackage{amsmath}
\usepackage{graphicx}
\usepackage{array}
\usepackage{booktabs}
\usepackage{xcolor}
\usepackage{colortbl}
\usepackage{pifont}
\usepackage{multirow}
\usepackage{placeins}
\usepackage{tikz}
\usetikzlibrary{arrows.meta,positioning}
\definecolor{Gray}{gray}{0.93}
\definecolor{deemph}{gray}{0.62}
\definecolor{mgreen}{rgb}{0.19,0.80,0.19}
\definecolor{selfblue}{RGB}{65,105,225} 
\definecolor{lightblue}{RGB}{220,230,255}
\definecolor{cellcolor}{HTML}{E3F2FD}
\definecolor{red}{HTML}{D32F2F}
\definecolor{magenta}{HTML}{D81B60}
\usepackage{fontawesome5}
\newlength\savewidth

\newcommand{\gc}[1]{\textcolor{deemph}{#1}}

\usepackage{enumitem}
\definecolor{ImproveBlue}{RGB}{245,249,253}
\usepackage[table]{xcolor}
\usepackage{makecell}

\usepackage[T1]{fontenc}
\usepackage[utf8]{inputenc}
\usepackage{microtype}
\usepackage{graphicx}

\newsavebox{\seqboxbuf}
\newenvironment{seqbox}[1]{%
  \setlength{\fboxsep}{0.5em}%
  \begin{lrbox}{\seqboxbuf}%
  \begin{minipage}{\dimexpr\columnwidth-2\fboxsep-2\fboxrule\relax}%
    \footnotesize
    {\sffamily\bfseries #1}\par
    \vspace{0.2em}\hrule\vspace{0.25em}%
    \ttfamily\raggedright
}{%
  \end{minipage}%
  \end{lrbox}%
  \par\addvspace{0.4em}\noindent\fbox{\usebox{\seqboxbuf}}\par\addvspace{0.4em}%
}
\newcommand{\exfield}[1]{{\normalfont\sffamily\bfseries #1}\ }

\acmYear{2027}
\acmConference[KDD '27]{Proceedings of the 33rd ACM SIGKDD Conference on
Knowledge Discovery and Data Mining}{August 2027}{TBD}

\begin{document}

\raggedbottom

\title{SeqLLM: Augmenting LLMs with Behavioral-Sequence Modeling for
High-Stakes Decisions at WeChat Pay}

\author{Guilin Li}
\authornote{Both authors contributed equally to this work.}
\affiliation{%
  \institution{WeChat Pay, Tencent}
  \country{China}
}

\author{Jiaxing Zhang}
\authornotemark[1]
\affiliation{%
  \institution{Shanghai Jiao Tong University}
  \country{China}
}

\author{Matthias Hwai Yong Tan}
\affiliation{%
  \institution{City University of Hong Kong}
  \country{Hong Kong SAR, China}
}

\author{Bo Wang}
\affiliation{%
  \institution{WeChat Pay, Tencent}
  \country{China}
}

\author{Weiran Huang}
\authornote{Corresponding author.}
\affiliation{%
  \institution{Shanghai Jiao Tong University}
  \country{China}
}

\renewcommand{\shortauthors}{Li et al.}

\begin{abstract}
Merchant risk control at large payment platforms screens tens of millions of
merchants per day, where errors are costly in both directions: a false positive
may harm a legitimate merchant, while a false negative leaves harmful activity
undetected. The hardest cases cannot be settled from either
modality alone; they require reading a merchant's textual profile jointly with
its long behavioral sequence. Large language models (LLMs) excel at the former
but cannot natively model behavioral sequences, and endowing them with this
ability typically erodes their language and reasoning skills through catastrophic
forgetting. We present SeqLLM, a framework that injects behavioral-sequence
modeling into a pretrained LLM while preserving its language ability, enabling
joint use of content and behavioral sequences. SeqLLM has three components: a
compact discrete behavior vocabulary that represents each event as native tokens;
a lightweight projector, trained with a two-stage alignment curriculum, that
grounds behavior tokens in the LLM's semantic space; and prefix-guided capability
injection, which learns sequence modeling through task-prefixed supervised
fine-tuning rather than continual pre-training. SeqLLM is fully deployed at WeChat Pay,
screening millions of merchants per day. Compared with
the production DeepSeek-based LLM baseline, it improves screening precision
from 92.0\% to 97.5\%; its pretrained behavior-token embeddings also lift Precision@Top-0.01\%
by 26.8 pp in a production fraud detector serving billion-scale
transaction traffic. Beyond payment behavior, SeqLLM achieves
state-of-the-art results against strong public baselines on open
recommendation benchmarks. On MovieLens and Amazon, it outperforms the strong User-LLM
baseline by up to 32\% relative Recall@5 while retaining markedly stronger
language ability. On the RecIF benchmark, SeqLLM improves
Pass@32 by 14.2\% over the full OneRec-8B pipeline, while using only one-fifth
of OneRec-8B's GPU-days. Our code is available at \url{https://github.com/125jx/SeqLLM.git}.
\end{abstract}

\maketitle

\section{Introduction}

At WeChat Pay, merchants include both offline businesses and online
payment-accepting entities, such as e-commerce platforms, mini-programs, and
apps. Merchant risk control screens millions of merchants each day for fraud, money laundering, and other illicit
fund flows~\cite{phua2010comprehensive,ngai2011financialfraud}. Criminals may
create online stores that appear legitimate and use them to collect illicit
payments. Errors are costly in both
directions: a false positive may harm a legitimate merchant, while a false
negative leaves harmful activity undetected, so each decision must be precise.

The hardest cases are still left to human experts. Consider a shop registered
under ``clothing'' that in fact sells jade via livestreams, luring buyers to pay
by a QR code sent in private chat: suspicious wording, but not conclusive on its
own. Its payments look ordinary too, until its behavioral sequence reveals that they
cluster at night, span many provinces, and target elderly buyers, with repeated
deleted bills.
Either clue alone is weak, and a pipeline that inspects text and behavioral
statistics separately misses the case; only reading the wording jointly with the
behavior reveals the fraud. Such expert review is accurate but slow, and does not
scale to tens of millions of merchants.

This need reflects a broader pattern. Entities on modern platforms are described
along two axes: their content, which captures what an entity is, and their
behavior, long sequences of timestamped actions that reveal how it acts over
time~\cite{caffagni2024mllm,kang2018self,wang2019sequential}. These axes drive
applications from recommendation~\cite{bobadilla2013recommender,rendle2010factorization,liu2020autofis,lin2025llm4rec}
and e-commerce search~\cite{vangysel2016productsearch,ai2017personalizedproductsearch,sarvi2020productsearch}
to risk control~\cite{li2026panther}, which increasingly demand reasoning over
both at once. Yet the two capabilities have long lived in disjoint model families:
large language models (LLMs) reason fluently over text but cannot natively consume
behavioral sequences~\cite{caffagni2024mllm,zhang2024mmllms}, while behavior models
capture temporal patterns but lack language, reasoning, and
explanation~\cite{kang2018self,li2026panther}. Recent work begins to bridge them,
injecting discretized behavior tokens into an
LLM~\cite{deng2025onerec,liu2025onerec,onereason2026} or contextualizing it with
a behavior-derived user embedding~\cite{ning2025user}. This progress has so
far centered on engagement-oriented recommendation, and its extension to
high-stakes decisions, where reliable outcomes require jointly leveraging
complementary evidence from text and behavioral sequences, remains largely
unexplored.

Equipping an LLM with behavioral-sequence modeling, however, is harder than it
looks. The obvious route---serializing a sequence as natural language---creates
long inputs and overloads word tokens with behavioral meanings. It also
represents temporal patterns in an embedding space designed for lexical
semantics, weakening the sequence signal.

Learning behavior with dedicated tokens avoids these pitfalls but raises a second
challenge: acquiring the new ability without eroding language competence.
OpenOneRec~\cite{zhou2025openonerec} acquires sequence capability through
recommendation-oriented continual pre-training (CPT), then relies on mixed-domain
training and general-ability distillation to counteract general-language
degradation. Yet this adapt-then-repair pipeline does not eliminate forgetting:
despite using roughly $4$M auxiliary language examples for only $156$K
behavioral sequences and a separate distillation stage, OpenOneRec still loses
general language ability. In our setting, sequences number in the tens of
millions ($\sim\!10^7$), a scale at which direct CPT drives a strong LLM's
general language ability to near-collapse (C-Eval
$0.78\!\rightarrow\!0.27$). General-language degradation therefore cannot be
reliably offset simply by scaling auxiliary language data.

\begin{figure*}[t]
  \centering
  \includegraphics[width=\textwidth]{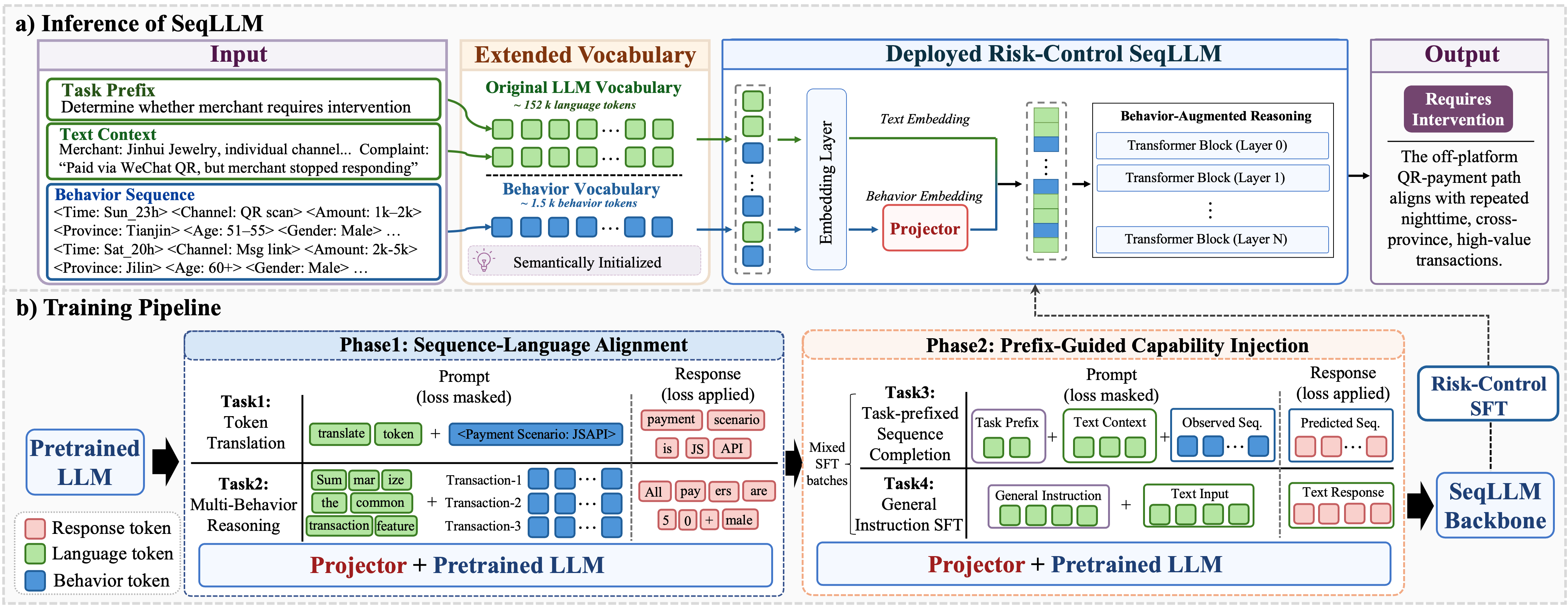}
  \caption{Overview of SeqLLM. \textbf{(a)}~At inference, projected behavior
  tokens are interleaved with the task prefix and textual context for joint
  text--behavior reasoning. \textbf{(b)}~Phase~1 grounds text-initialized
  behavior tokens through token translation and multi-event reasoning. Phase~2
  jointly mixes task-prefixed sequence completion with general-instruction SFT,
  applying loss only to response tokens. This injects sequence modeling without
  behavior-oriented CPT or a separate capability-recovery stage. The resulting
  behavior-capable SeqLLM backbone is adapted through risk-control SFT for
  deployment.}
  \label{fig:overview}
\end{figure*}

To meet both challenges, we propose \textbf{SeqLLM}, a framework that can be
applied to pretrained  LLMs, endowing them with native
behavioral-sequence modeling ability while preserving their language ability
and supporting downstream tasks that jointly use textual and behavioral
information. SeqLLM rests on three components
(Figure~\ref{fig:overview}).
First, a discrete behavior vocabulary encodes each event as field-level tokens,
compressing it from dozens of tokens to about nine on average (roughly
$6\times$) while avoiding collisions with the original word vocabulary.
Second, inspired by visual--language alignment~\cite{liu2023llava}, a lightweight
behavior projector with text-grounded initialization aligns these tokens into the
LLM's semantic space through a two-stage, translation-then-reasoning curriculum,
so the model first reads behavior tokens and then reasons over them. Third, and
central to our findings, prefix-guided capability injection learns sequence
modeling through instruction-conditioned SFT rather than CPT: by recasting
next-event prediction as conditional generation under a task prefix, sequence
learning is confined to the parameter pathways the prefix activates, avoiding the
broad general-language degradation caused by behavior-oriented CPT.

We evaluate SeqLLM in two large-scale business scenarios at WeChat Pay and on
three public recommendation benchmarks. The results demonstrate
three main findings. First, sequence capability can be injected at scale
without eroding language ability: trained on approximately $20$M unlabeled
merchant behavior sequences,
prefix-guided SFT matches CPT on next-event prediction (HR@10 $0.80$) while
 retaining a C-Eval score of $0.78$, versus $0.27$ under CPT. Second, the
 resulting model delivers measurable production impact at WeChat Pay. SeqLLM is
 deployed in two complementary roles: as an end-to-end merchant screening model
that processes millions of merchants per day, and as a behavior-embedding
provider for an existing fraud detector. In the first role, a three-month
shadow evaluation confirms $97.5\%$ of flagged merchants as risky versus
$92.0\%$ for the production baseline, while the post-launch appeal rate drops
from $12\%$ to $\sim2\%$ with no exonerations. In the second role, an online
A/B test shows that initializing the fraud detector's behavior-token
embeddings with those learned by SeqLLM yields gains of 26.8 pp in
Precision@Top-0.01\% and 33.1 pp in Recall@Top-1\%---the largest improvement
across historical iterations of this online fraud-detection model.
Third, the methodology generalizes beyond payment behavior to public
recommendation benchmarks. On MovieLens and Amazon, SeqLLM improves Recall@5
over User-LLM~\cite{ning2025user} by up to $32\%$. On RecIF, SeqLLM+RL improves
Pass@32 by $14.2\%$ over OpenOneRec's full OneRec-8B
pipeline~\cite{zhou2025openonerec} while using $4.8\times$ fewer GPU-days and
no separate distillation stage. To test behavior-token understanding beyond
ranking accuracy, we construct RecProbe, a four-task evaluation suite; SeqLLM
outperforms OneRec-8B on all four tasks, raising Video--Topic Matching
accuracy from $0.465$ to $0.745$.

Our contributions are fourfold:
\begin{enumerate}[label=(\arabic*), leftmargin=1.5em, labelsep=0.4em,
  itemsep=1pt, topsep=2pt, parsep=0pt, partopsep=0pt]
  \item We formulate \emph{joint text--behavior modeling} for high-stakes
  decisions: a regime where risk emerges only from the interaction between
  textual evidence and long event streams, and is therefore systematically
  missed by text-only or behavior-only pipelines. Controlled modality
  experiments confirm the value of joint modeling
  (Table~\ref{tab:v2-real-world}).
  \item We introduce SeqLLM, a general framework that endows a pretrained LLM
  with native behavioral-sequence modeling through a compact field-level vocabulary,
  a text-grounded residual projector with a translation-to-reasoning curriculum,
  and prefix-guided capability injection.
  \item We establish prefix-guided SFT as a scalable alternative to
  adapt-then-repair CPT. It matches CPT on sequence modeling while preserving
  language ability, without a separate general-ability distillation stage.
  \item We demonstrate impact at both industrial and public scale. SeqLLM is fully
  deployed at WeChat Pay in two production systems for merchant screening and
  fraud detection. On public benchmarks, it substantially outperforms User-LLM
  and OpenOneRec's full pipeline in both recommendation accuracy and
  behavior-token semantic understanding, while preserving general language
  ability and using up to $4.8\times$ fewer GPU-days.
\end{enumerate}

\section{Related Work}
\label{sec:related}

\paragraph{Behavioral sequence modeling.}
Behavioral sequence models learn temporal patterns from interaction
histories~\cite{wang2019sequential}, from recurrent and Transformer
architectures such as GRU4Rec~\cite{hidasi2015session},
SASRec~\cite{kang2018self}, and BERT4Rec~\cite{sun2019bert4rec} to large
generative systems like HSTU~\cite{zhai2024actions} and
TIGER~\cite{rajput2023recommender}. These models are strong at behavioral prediction
but do not natively support free-form language reasoning over textual evidence.

\paragraph{LLMs for behavioral sequences.}
Recent work brings behavior into LLMs~\cite{lin2025llm4rec}. Text-based systems
such as P5~\cite{geng2022recommendation} serialize interactions as language, providing a
natural interface at the cost of long contexts and overloaded word tokens.
Generative recommenders instead represent items with dedicated codes:
OneRec~\cite{deng2025onerec} unifies retrieval and ranking,
OpenOneRec~\cite{zhou2025openonerec} grounds itemic tokens in language, and
OneRec-Think~\cite{liu2025onerec} and OneReason~\cite{onereason2026} develop
explicit reasoning over itemic histories. A complementary approach,
User-LLM~\cite{ning2025user}, conditions an LLM on a compact user vector
injected via cross-attention.

\paragraph{Positioning SeqLLM}
These methods show LLMs can model behavioral histories, but our setting poses a
different task: high-stakes risk control must jointly weigh heterogeneous textual
evidence, such as merchant profiles and complaints, against field-rich
transaction events to produce an auditable decision. OpenOneRec is our closest
generative counterpart: it aligns newly added itemic embeddings directly in the
expanded embedding table without an explicit modality projector, then acquires
sequence modeling through full-parameter co-pretraining, with mixed-domain
data and general-ability distillation repairing the resulting general-language
degradation.
Yet this adapt-then-repair pipeline does not eliminate forgetting: even with
roughly $4$M auxiliary language examples for only $156$K behavioral sequences
plus a separate distillation stage, OpenOneRec still loses general language
ability, and the gap widens as the behavioral corpus grows. SeqLLM instead
uses a text-grounded projector and prefix-guided SFT to inject sequence
capability without a recovery stage, breaking this coupling. User-LLM avoids drift only in its frozen-backbone (Enc) variant;
its default Full strategy finetunes the backbone, and either way the entire
history is compressed into one vector. We report the matched quantitative
comparison in Sections~\ref{sec:v2-injection} and~\ref{sec:v2-generalization}.

\paragraph{Risk detection.}
Fraud and risk detection has progressed from rules and statistical scoring to
learned models over transaction graphs and behavioral
sequences~\cite{phua2010comprehensive,ngai2011financialfraud}, e.g.,
PANTHER~\cite{li2026panther} pretrains on transaction histories and transfers to
downstream risk tasks. Existing systems nevertheless tend to score textual and
behavioral signals in separate pipelines. SeqLLM instead reasons over both
within one model, targeting risks that become conclusive only through their
interaction.

\section{Methodology}
\label{sec:methodology}

SeqLLM injects behavioral-sequence modeling into a pretrained LLM through three
components, each addressing one challenge of joint text--behavior modeling: how to
represent behavior compactly, how to make the LLM understand it, and how to
inject sequence modeling without eroding language ability. First, a
discrete behavior vocabulary encodes each event as compact field-level
tokens (Section~\ref{sec:behavioral-vocabulary}). Second, a lightweight
behavior projector aligns these tokens into the LLM's semantic space via a
translation-then-reasoning curriculum (Section~\ref{sec:behavioral-projector}).
Third, prefix-guided capability injection acquires sequence modeling
through task-prefixed SFT rather than CPT, preserving the backbone's language
ability (Section~\ref{sec:prefix-guided-capability-injection}).
\subsection{Problem Formulation}
\label{sec:problem-formulation}

We consider an entity $e$---a merchant, user, or item---with textual content
$\mathcal{T}_e$ describing what it is and a timestamped behavioral sequence
$\mathcal{S}_e=\langle s_1,\ldots,s_{L_e}\rangle$ describing how it acts. Each
event $s_t\in\mathcal{E}$ contains fields such as time, amount, channel, and
event status (Section~\ref{sec:behavioral-vocabulary}). Given a natural-language
instruction $\mathcal{I}$, a single model $\Theta$ jointly uses both signals to
model the task-specific output $\mathbf{y}_e$ as
\begin{equation}
  p_{\Theta}(\mathbf{y}_e \mid \mathcal{I}, \mathcal{T}_e, \mathcal{S}_e).
  \label{eq:seqllm-objective}
\end{equation}
Depending on $\mathcal{I}$, $\mathbf{y}_e$ may be a risk decision with a
natural-language rationale or the next event $s_{L_e+1}$. In high-stakes
screening, neither modality may suffice alone. Figure~\ref{fig:overview}(a)
illustrates the jade-shop case: a category mismatch and private-QR complaint
become conclusive only when read with nighttime, cross-province, high-value
payments and a deleted bill. SeqLLM must therefore consume long event streams
without compromising the LLM's language and reasoning ability.

\subsection{Behavioral Vocabulary}
\label{sec:behavioral-vocabulary}

The first design choice is how to represent each event. Instead of writing it
out as text, we split the event into its fields---time, amount, event status,
and so on---and turn each field value into its own dedicated token. For
example, one payment is encoded as follows:
\par\smallskip
{\small\ttfamily\raggedright
<Time:Thu\_00h> <Scene:App pay> <Channel:Scan QR>\\
<Amount:200-500 CNY> \ldots\ <Status:Success>\par}
\smallskip\noindent
Each angle-bracketed unit is one token. Formally, a field $f_i$ takes discretized values
$V_i$ (e.g., log-scale amount buckets or an event-status code); each value
$v\in V_i$ maps to a token $\langle f_i\!:\!v\rangle$, and the behavioral
vocabulary is their union $\mathcal{V}_{b}=\bigcup_i \{\langle f_i\!:\!v\rangle :
v\in V_i\}$. An event $s_t$ is the concatenation of its field
tokens and a sequence $\mathcal{S}_e$ the concatenation of its events; the event
space $\mathcal{E}$ collects all such field-token strings. These tokens occupy a
disjoint sub-range of the embedding table, insulating behavior from the word
vocabulary while reusing the same backbone.

Field-level factorization offers three benefits. First, it is compact: an
additive vocabulary of $\sum_i |V_i|$ entries represents the
$\prod_i |V_i|$ combinatorial event space using about $9$ field tokens per
event on average---each event type encodes its applicable field subset
(Appendix~\ref{sec:appendix-wechat-data})---far fewer than serialized text and
without reusing word tokens. Second, it
supports compositional generalization: unlike whole-event IDs, as used by
PANTHER~\cite{li2026panther}, events differing in one field share all remaining
tokens, so novel combinations remain expressible and decisions can be attributed
to individual fields. Third, the tokens are semantically readable: explicit
field--value names enable text-based initialization and traceable decisions,
whereas RQ-VAE codes carry no intrinsic textual meaning and require a separate
quantizer (Section~\ref{sec:behavioral-projector}).

\subsection{Behavior Projector and Sequence-Language Alignment}
\label{sec:behavioral-projector}

The new field tokens enter with no trained embedding. A model can learn to
predict them as a bare sequence without any grounding, but understanding and
reasoning over them requires aligning them to the LLM's semantic space.
OpenOneRec~\cite{zhou2025openonerec}, for instance, trains newly added itemic
embeddings through item--text alignment before full-parameter co-pretraining,
without an explicit projection interface.

SeqLLM instead casts grounding as an interface-and-curriculum problem:
a lightweight behavior projector supplies each token a dedicated route into the
LLM, and a two-stage translation-then-reasoning curriculum first anchors that
route in language, then opens it to reasoning (Stages 1--2 below). Alignment
thereby advances from understanding individual tokens to reasoning over them.

\noindent\textbf{\textit{Behavior projector.}}
Every newly added behavior token reaches the LLM through a shared projector
\begin{equation}
  g_\psi(\mathbf{e})=\mathbf{e}+\text{MLP}_\psi(\mathbf{e}),
\end{equation}
a two-layer MLP (Linear--GELU--Linear) with a residual (skip) connection whose
final linear layer is zero-initialized, so $g_\psi$ starts as the identity and
learns only a small correction to each input embedding $\mathbf{e}$. Applying
the same transformation to every behavior token imposes a shared alignment
constraint: related field tokens are mapped consistently into the LLM's semantic
space, allowing the backbone to compose them across fields and events. Directly
tuning each token embedding provides no such cross-token constraint.
This distinction is empirical, not merely architectural. In a controlled
industrial ablation, the w/o-projector variant learns to translate individual
tokens but fails to transfer their meanings to multi-event reasoning and
downstream decisions; the shared projector closes this gap
(Appendix~\ref{sec:supp-projector-ablation},
Table~\ref{tab:supp-projector-ablation}). The same pattern appears on RecIF,
where SeqLLM outperforms OpenOneRec's direct-alignment pipeline on all four
semantic-understanding and preference-reasoning probes
(Table~\ref{tab:v2-rec_understanding}).

Because our field tokens are readable, we ground each in the text it denotes
rather than initializing at random. Let $Z_v$ be the backbone tokenizer's
segmentation of token $v$'s readable text (e.g., \texttt{<Amount:200-500 CNY>}
$\to$ \texttt{Amount 200-500 CNY}); we mean-pool the corresponding embeddings and
rescale the result to the backbone's embedding statistics,
\begin{equation}
  \bar{\mathbf{e}}_v=\frac{1}{|Z_v|}\sum_{z\in Z_v}\mathbf{E}[z],
  \qquad
  \mathbf{e}_v^{(0)}=\boldsymbol{\mu}+\boldsymbol{\sigma}\odot
  \frac{\bar{\mathbf{e}}_v-\hat{\boldsymbol{\mu}}}{\hat{\boldsymbol{\sigma}}},
\end{equation}
where $\mathbf{E}$ is the backbone embedding table, $(\boldsymbol{\mu},
\boldsymbol{\sigma})$ its per-dimension mean and standard deviation, and
$(\hat{\boldsymbol{\mu}},\hat{\boldsymbol{\sigma}})$ those of the pooled vectors
$\{\bar{\mathbf{e}}_v\}$. This rescaling step is essential: without it, the
pooled vectors fall outside the backbone's embedding distribution and disrupt
the model; rescaling them to match its mean and variance places each token in
the same region as its own descriptive words. 

Recall from Section~\ref{sec:behavioral-vocabulary} that event $s_t$ concatenates
field tokens; we write $\Phi(s_t)$ for its token string ($\Phi(\mathcal{S}_e)$ for
a multi-event window) and train the projector in two stages.

\noindent\textbf{\textit{Stage 1: Translation.}}
We first optimize a translation objective that reconstructs the natural-language
reading of behavior tokens, training the projector $\psi$ jointly with the LLM.
Each instance pairs an instruction $\mathcal{I}$, an input $x=\Phi(s_t)$ (or a
short window), and a field-wise target $y$; an illustrative example from our
translation corpus is:
\begin{seqbox}{Stage 1 example: translation}
\exfield{Instruction:} Translate these behavior tokens.\par\smallskip
\exfield{Input:} <Time:Sun\_14h> <Channel:Msg link> <Status:Success>\par\smallskip
\exfield{Output:} Sunday at 14:00; successful message-link payment.
\end{seqbox}
\noindent
The bridge loss is
\begin{equation}
  \mathcal{L}_{\text{bridge}}=-\!\!\!\sum_{(\mathcal{I},x,y)\in\mathcal{D}_{\text{trans}}}\!\!\!
  \log p_{\Theta,\psi}(y\mid\mathcal{I},x),
\end{equation}
where embeddings of tokens in $x$ pass through $g_\psi$ before entering the LLM.
This objective anchors each field token to its meaning, grounding the vocabulary
before the model is asked to reason over it.

\noindent\textbf{\textit{Stage 2: Reasoning.}}
We then continue with SFT on multi-event reasoning queries over
real merchant sequences. Each instance provides $\mathcal{I}$, a window
$x=\Phi(\mathcal{S}_e)$ of tokenized transactions, and a free-text answer that
aggregates or compares fields---customer profiling, anomaly spotting, or risk
commentary. A representative example is:
\begin{seqbox}{Stage 2 example: reasoning}
\exfield{Instruction:} What pattern is shared by these transactions?\par\smallskip
\exfield{Input:} Tx1: <Time:Tue\_15h> <Channel:Scan QR>;
Tx2: <Time:Tue\_16h> <Channel:Scan QR>\par\smallskip
\exfield{Output:} Both are Tuesday-afternoon QR payments.
\end{seqbox}
\noindent
Together, translation grounds each new token in the LLM's semantic space,
while reasoning alignment enables the model to interpret and use this new
vocabulary across contexts, preparing it for joint text--behavior decisions in
Section~\ref{sec:prefix-guided-capability-injection}. We present the
two as stages for clarity; since translation is lightweight, their data can also
be mixed into a single pass with equivalent effect.

\subsection{Prefix-Guided Capability Injection}
\label{sec:prefix-guided-capability-injection}

Token grounding teaches the model what each behavior token denotes, but not
what sequences of such tokens reveal about an entity. Platform-specific
regularities---such as spending rhythms, amount transitions, channel
preferences, and deviations from routine---are largely absent from language
pretraining. Learning them requires modeling event co-occurrence and temporal
evolution. Future-event prediction provides this supervision: generating a
continuation requires the model to infer regularities from the observed
history. Standard CPT applies next-token loss at every position in the behavior
stream. Although this learns sequence structure, it also updates the backbone
broadly toward behavior-token prediction, risking interference with pretrained
language knowledge.

Our key idea is to retain future-event prediction while turning it from a
default modeling objective into an instruction-conditioned capability. The
instruction acts as a task condition: behavior continuation is optimized only
when this condition is present, rather than being imposed on every input. For
each sequence, we choose a cut $k$ at about $70\%$ of its length. The first part,
$c=\Phi(s_1,\ldots,s_k)$, is provided as input under a natural-language
instruction $\mathcal{I}$, while the remaining events,
$y=\Phi(s_{k+1},\ldots,s_{L_e})$, form the answer. Loss is computed only on this
answer. Let $b_{1:N}=\Phi(\mathcal{S}_e)$ denote the resulting behavior-token
sequence and $m$ the token boundary induced by the event cutoff $k$.
Figure~\ref{fig:overview}(b) illustrates the construction; at the token level,
the two objectives differ in where the loss is applied:
\begin{equation}
\begin{aligned}
\mathcal{L}_{\mathrm{CPT}}
  &= -\sum_{t=1}^{N}\log p_{\Theta}(b_t\mid b_{<t}),\\
\mathcal{L}_{\mathrm{Prefix}}
  &= -\sum_{t=m+1}^{N}\log
     p_{\Theta,\psi}(b_t\mid \mathcal{I},b_{<t}).
\end{aligned}
\label{eq:cpt-prefix-token-loss}
\end{equation}

\textbf{\textit{Why this injects sequence ability.}}
The sequence supervision is retained rather than removed. The remaining
$\sim30\%$ of events form the response and receive the same autoregressive
next-token loss used by CPT. Predicting each response token from the preceding
history directly trains temporal dependencies, both from the observed prefix
$s_{1{:}k}$ to the target suffix $s_{k+1{:}L_e}$ and within the suffix itself.
Prefix masking only removes loss on the observed $70\%$; it leaves the
future-event targets---and hence the core sequence-learning signal of CPT---
intact.

\noindent\textbf{\textit{Why this mitigates forgetting.}}
CPT treats behavior prediction as an unconditional objective and applies loss
at almost every position, causing the behavior corpus to update the backbone
broadly. Prefix-guided SFT instead conditions behavior prediction on an
explicit instruction and applies supervision only to the response suffix.
This confines the behavior objective to a specific task context, reducing
interference with pretrained language capabilities while preserving the
future-prediction signal. Consistent with this interpretation, prefix-guided
SFT produces smaller and more uniform per-layer weight changes than CPT
(Figure~\ref{fig:v2-deltaw}). It matches sample-aligned CPT on sequence
modeling (HR@10 $0.801$ vs.\ $0.802$) while preserving substantially more
language ability (C-Eval $0.783$ vs.\ $0.271$;
Table~\ref{tab:v2-main_results}).

We optimize the backbone and projector with response-only loss:
\begin{equation}
  \mathcal{L}_{\text{inject}}=-\!\!\!\sum_{(\mathcal{I},c,y)\in\mathcal{D}_{\text{inj}}}\!\!\!
  \log p_{\Theta,\psi}(y\mid \mathcal{I}, c),
  \label{eq:inject}
\end{equation}
For sequence examples, all behavior tokens pass through $g_\psi$.
$\mathcal{D}_{\text{inj}}$ also includes general instruction examples, whose
response loss preserves language capabilities during sequence injection,
avoiding a separate capability-recovery stage.

\section{Experiments \& Results}

We evaluate SeqLLM on large-scale payment data from WeChat Pay and on public
recommendation benchmarks, complemented by evidence from production deployments.
Our experiments address four questions:
\begin{itemize}[leftmargin=1.4em, itemsep=2pt, topsep=3pt, parsep=0pt]
  \item \textbf{RQ1}: Can SeqLLM enable a pretrained LLM to model behavioral
  sequences while retaining its general language capabilities?
  \item \textbf{RQ2}: Does joint text--behavior modeling improve merchant risk
  screening over text-only and behavior-only modeling?
  \item \textbf{RQ3}: Does deploying SeqLLM yield measurable production gains
  in merchant screening and fraud detection?
  \item \textbf{RQ4}: Does SeqLLM generalize beyond payment behavior to diverse
  recommendation domains, and how does it compare with strong public
  sequence--language baselines in recommendation performance, semantic
  understanding, and language retention?
\end{itemize}
We answer \textbf{RQ1} through a controlled comparison with CPT on WeChat Pay
data, \textbf{RQ2} through a controlled modality comparison, and \textbf{RQ3}
through two production deployments. For \textbf{RQ4}, we compare with User-LLM
on MovieLens and Amazon and with OpenOneRec's OneRec-8B on RecIF, including
semantic and preference probes.

\subsection{Experiment Setup}
\label{sec:v2-setup}

\paragraph{Settings.} We consider three settings:
\textbf{(i)} behavioral-sequence modeling on industrial payment data and two
downstream deployments at WeChat Pay: merchant screening and fraud detection;
\textbf{(ii)} next-item prediction, preference inference, and review generation
on MovieLens-20M and Amazon Reviews; and
\textbf{(iii)} sequential recommendation and item understanding on the RecIF
benchmark, where we compare with the unified sequence--language model
OneRec-8B released by OpenOneRec~\cite{zhou2025openonerec}.

\paragraph{Backbone \& compute.} Unless stated otherwise, SeqLLM uses Qwen3-8B
as its pretrained backbone. We report training cost in GPU-days; full hardware
and training configurations are provided in Appendix~\ref{sec:hyperparameter}.
Code for all public-benchmark experiments, including the RecProbe construction
scripts, will be open-sourced.

\paragraph{Datasets.} We evaluate on industrial data from two WeChat Pay
production scenarios and three public recommendation datasets; full statistics and preprocessing are provided
in Appendix~\ref{sec:appendix-data}.
\begin{itemize}[leftmargin=1.4em, itemsep=1pt, topsep=2pt, parsep=0pt]
  \item \textbf{WeChat Pay.} Each merchant includes a profile, complaint text,
  and transaction sequence. We inject sequence capability using $\sim$20M
  unlabeled sequences without risk-positive merchants, then perform risk SFT on
  4.06M labeled merchants from a preceding multi-month window. Evaluation
  covers a subsequent 30-day window ($\sim$0.99M merchants/day); features are
  cut at scoring time.
  After scoring each merchant, we wait 30 days to collect subsequent evidence
  of risk before assigning its final label. Histories are capped at $1{,}000$
  events and inputs at $10{,}000$ tokens.
  \item \textbf{MovieLens-20M \& Amazon Reviews.} We evaluate next-item
  prediction, preference classification, and review generation following the
  User-LLM protocol~\cite{ning2025user}; see
  Appendix~\ref{sec:appendix-public-data}.
  \item \textbf{RecIF.} This OpenOneRec benchmark contains 96M interactions from
  160K users and evaluates sequential recommendation and item understanding under
  its official protocol~\cite{zhou2025openonerec}; see
  Appendix~\ref{sec:appendix-recif-data}.
\end{itemize}

\paragraph{Metrics.} For next-transaction prediction, HR@10 is the fraction of
examples whose ground-truth next event appears in the top 10, following
PANTHER~\cite{li2026panther}. Recall@5 analogously tests whether the held-out next
item appears in the top 5. For merchant risk, we report the risky fraction among
the highest-scored $r\%$ of merchants ($r\in\{1,0.1,0.01\}$).
Following OpenOneRec~\cite{zhou2025openonerec}, RecIF
reports whether the target appears among $k$ generated candidates (Pass@$k$) and
the fraction of relevant items retrieved (Recall@$k$). Other task metrics are
Accuracy, NDCG@3, and ROUGE; general language ability is measured by
MMLU~\cite{hendrycks2020measuring}, C-Eval~\cite{huang2023c}, and
AGIEval~\cite{zhong2024agieval}. Higher is better throughout; implementation
details are in Appendix~\ref{sec:appendix-evaluate}.

\FloatBarrier
\subsection{Sequence Modeling without Catastrophic Forgetting (RQ1)}
\label{sec:v2-injection}

This section compares prefix-guided SFT with CPT to show that the former
injects behavioral-sequence capability without catastrophic forgetting.
Sample-aligned CPT matches our behavioral examples and updates, while
gradient-aligned CPT matches our loss-bearing behavior tokens. We test both
under no behavior-token alignment and the full alignment stack. All variants
share the backbone, corpora, and optimizer; CPT additionally receives both
raw-text and chat-format language replay, giving it strictly more language
supervision (Appendix~\ref{sec:appendix-cpt}).
PANTHER~\cite{li2026panther} and zero-shot Qwen3-8B provide sequence-only and
language-only references.
\begin{table}[!htbp]
    \centering
    \footnotesize
    \caption{Sequence modeling and language retention for prefix-guided SFT and
    matched CPT controls under two alignment settings. ``No alignment stack''
    disables semantic initialization, the projector $g_\psi$, and the
    translation$\to$reasoning curriculum; ``full alignment stack'' enables all
    three. Component-level ablations: Table~\ref{tab:v2-public},
    Table~\ref{tab:supp-projector-ablation}.}
    \label{tab:v2-main_results}
    \renewcommand{\arraystretch}{1.05}
    \setlength{\tabcolsep}{3pt}
    \begin{tabular}{@{}lcccc@{}}
    \toprule
    \textbf{Method} & \textbf{HR@10} & \textbf{C-Eval}
    & \textbf{MMLU}
    & \textbf{AGIEval} \\
    \midrule
    PANTHER~\cite{li2026panther} {\scriptsize (sequence-only)}
    & 0.680 & -- & -- & -- \\
    Qwen3-8B~\cite{yang2025qwen3} {\scriptsize (zero-shot)}
    & 0.312 & 0.779 & \textbf{0.769} & 0.636 \\
    \midrule
    Qwen3-8B + CPT  {\scriptsize (sample-aligned)}
    & 0.802 & 0.271 & 0.264 & 0.324 \\
    Qwen3-8B + CPT  {\scriptsize (gradient-aligned)}
    & 0.784 & 0.435 & 0.467 & 0.379 \\
    Prefix-guided SFT {\scriptsize (ours)}
    & 0.801 & 0.783 & 0.745 & 0.640 \\
    \midrule
    Qwen3-8B + CPT + align  {\scriptsize (sample-aligned)}
    & 0.804 & 0.293 & 0.298 & 0.367 \\
    Qwen3-8B + CPT + align  {\scriptsize (gradient-aligned)}
    & 0.793 & 0.456 & 0.489 & 0.377 \\
    \rowcolor{ImproveBlue}
    \textbf{SeqLLM (ours)}
    & \textbf{0.806} & \textbf{0.789} & 0.765 & \textbf{0.643} \\
    \bottomrule
    \end{tabular}
    \end{table}
Table~\ref{tab:v2-main_results} shows that SeqLLM matches CPT on sequence
modeling (HR@10 $0.806$ vs.\ $0.804$) while preserving the original language
ability (C-Eval $0.789$ vs.\ $0.293$). Two further controls rule out
alternative explanations: the gradient-aligned CPT row matches the
cumulative number of loss-bearing behavior tokens yet still collapses
language ability, so the retention advantage is not explained by
supervision volume; and the same pattern holds in the no-alignment setting,
so it is not contingent on the alignment stack. The high-replay-ratio
OpenOneRec comparison---where OneRec-8B also uses continual pre-training---reaches the same conclusion
(Section~\ref{sec:v2-generalization}).

\begin{figure}[htbp]
    \centering
    \includegraphics[width=0.78\linewidth]{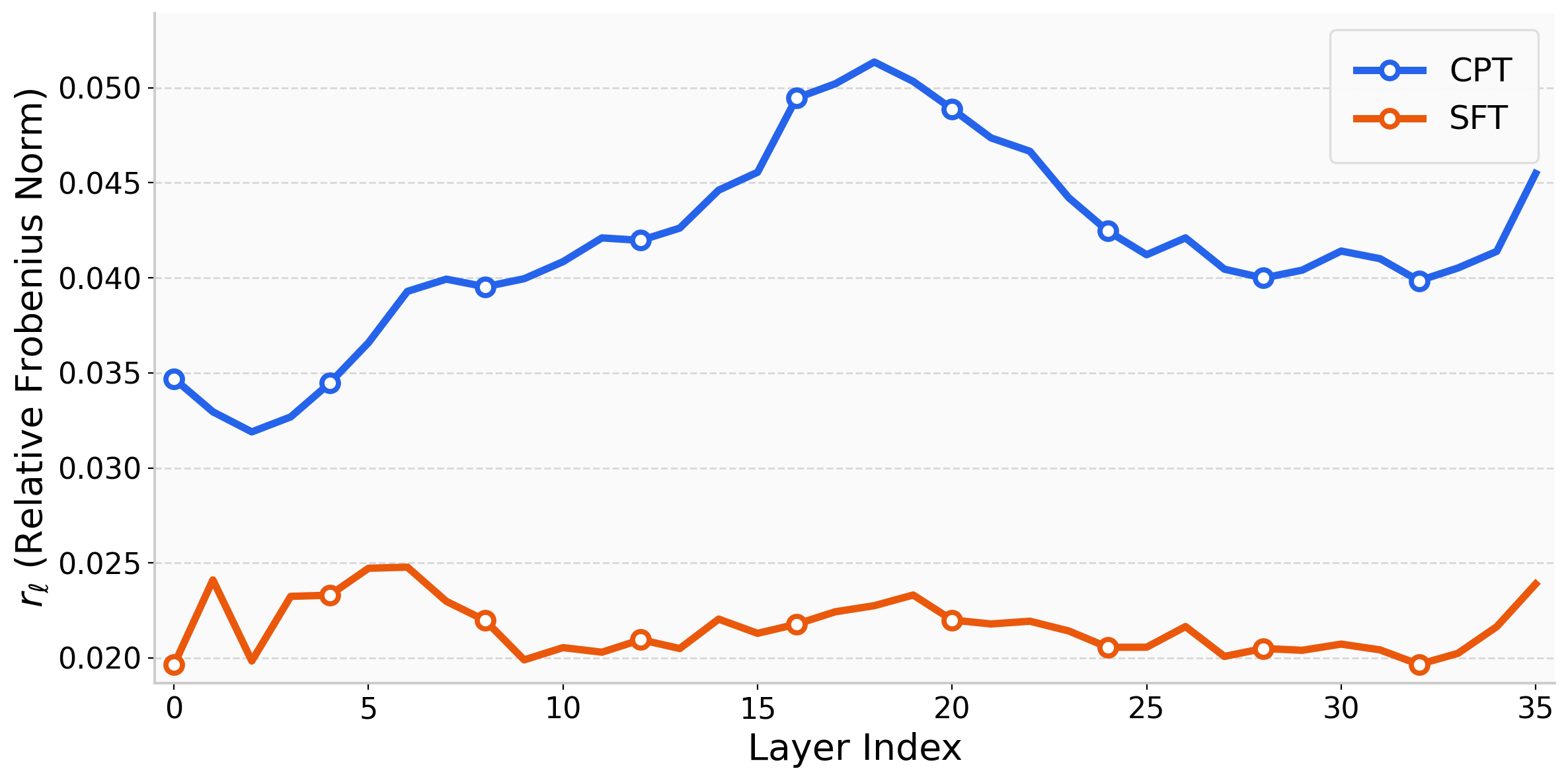}
    \caption{Per-layer relative weight change after sample-aligned CPT and
    prefix-guided SFT training from the same Qwen3-8B backbone.}
    \label{fig:v2-deltaw}
\end{figure}

\paragraph{Parameter-level evidence: prefix-guided SFT better preserves the backbone.}
With the same behavioral and replay corpora, sample-aligned CPT produces much
larger per-layer changes, especially in the middle Transformer blocks, whereas
prefix-guided SFT produces smaller, more uniform updates
(Figure~\ref{fig:v2-deltaw}). These smaller updates provide parameter-level
evidence for why prefix-guided SFT better preserves general language ability.

\subsection{Multimodal Merchant Risk Modeling (RQ2)}
\label{sec:v2-multimodal-risk}
To answer \textbf{RQ2}, we compare controlled text-only, behavior-only, and
joint variants that use the same SeqLLM architecture, training data, and
supervision. This isolates whether text and behavioral sequences provide
complementary evidence for merchant risk screening.

\paragraph{Task and downstream adaptation.}
We first train the SeqLLM backbone through sequence--language alignment and
prefix-guided sequence-capability injection
(Sections~\ref{sec:behavioral-projector}
and~\ref{sec:prefix-guided-capability-injection}), then fine-tune it on the
labeled merchant risk-SFT corpus. Each example asks the model whether risk
control is required, with the answer beginning with either the ``control''
or ``no control'' label. For ranking, rather than using only the generated
answer, we score each merchant by the normalized probability of the positive
label token at the first response position,
$s=\exp(\ell^{+})/[\exp(\ell^{+})+\exp(\ell^{-})]$, and report precision among
the highest-scored merchants. The controlled modality setup is detailed in
Appendix~\ref{sec:appendix-baselines}.

\begin{table}[htbp]
\centering
\small
\caption{Offline merchant-risk precision under controlled input modalities.}
\label{tab:v2-real-world}
\renewcommand{\arraystretch}{1.1}
\setlength{\tabcolsep}{4pt}
\begin{tabular}{@{}lccc@{}}
\toprule
\textbf{Input modality}
& \textbf{P@Top-1\%}
& \textbf{P@Top-0.1\%}
& \textbf{P@Top-0.01\%} \\
\midrule
Behavior only        & 9.0\%  & 20.1\% & 70.0\% \\
Text only            & 23.0\% & 70.1\% & 88.0\% \\
\rowcolor{ImproveBlue}
Text + Behavior      & \textbf{32.6\%} & \textbf{79.2\%}
                     & \textbf{97.0\%} \\
\bottomrule
\end{tabular}
\end{table}

Table~\ref{tab:v2-real-world} isolates the contribution of joint modeling by
holding the model and supervision fixed while varying only the input. Behavior
alone misses merchant context, while text alone misses temporal transaction
patterns; combining both raises precision substantially at every cutoff,
reaching $97.0\%$ at Top-$0.01\%$. This confirms that the two modalities
provide complementary evidence for merchant risk screening. 

\subsection{Two Production Deployments at WeChat Pay (RQ3)}
\label{sec:v2-deployment}
To answer \textbf{RQ3}, we evaluate SeqLLM in two complementary production
roles. The first uses the joint model directly for merchant risk screening; the
second uses SeqLLM-pretrained behavior-token embeddings to initialize the
embedding layer for behavior-sequence events in a downstream fraud detector.

\paragraph{Deployment I: merchant risk screening.}
SeqLLM is deployed as a 0.6B--8B cascade: a 0.6B scanner scores approximately
50 million active merchants daily and retrieves a fixed candidate set, which
the 8B model ranks for risk-control action; the full service runs on 128 GPUs.
The production baseline is a DeepSeek-based LLM adapted through SFT and RL,
using merchant profiles and complaints but no behavioral sequence.

Before launch, we ran a three-month matched prospective shadow evaluation of
three systems---the baseline, DeepSeek with text-serialized behavior (same
backbone and recipe, differing only in input), and SeqLLM's 8B ranking stage.
Each day, all three ranked the same neutral candidate pool from an upstream
pre-screen independent of all evaluated systems, and returned the same number
of top candidates; shadow outputs did not affect review, enforcement, or
labels. A candidate is labeled positive if confirmed within 30 days either by
the existing production risk system---an ensemble of expert strategies operated
independently of all three evaluated scorers, including the DeepSeek
baseline---or through subsequently confirmed user-reported harm.

\begin{table}[htbp]
\centering
\scriptsize
\caption{Merchant screening: three-month shadow evaluation and subsequent
production. All metrics are weekly-averaged; appeal and exoneration rates are
operational indicators collected after the two systems are officially
deployed.}
\label{tab:v2-online-deployment}
\renewcommand{\arraystretch}{1.05}
\setlength{\tabcolsep}{2.5pt}
\resizebox{\columnwidth}{!}{%
\begin{tabular}{@{}lccc@{}}
\toprule
\textbf{System (input)} &
\makecell{\textbf{Risk precision}\\\textbf{(paired shadow; $\uparrow$)}} &
\makecell{\textbf{Appeal rate}\\{\tiny post-launch, among actioned; $\downarrow$}} &
\makecell{\textbf{Exoneration rate}\\{\tiny post-launch, among appeals; $\downarrow$}} \\
\midrule
DeepSeek baseline (text only) & $92.0\%$ & $12\%$ & $8\%$ \\
DeepSeek + serialization & $83.0\%$ & -- & -- \\
\rowcolor{ImproveBlue}
SeqLLM (text + behavior) & $97.5\%$ & $\sim2\%$ & $0\%$ \\
\midrule
\textit{Improvement} & $+5.5$ pp & $10$ pp reduction & $8$ pp reduction \\
\bottomrule
\end{tabular}%
}
\end{table}

As shown in Table~\ref{tab:v2-online-deployment}, SeqLLM improves
weekly-averaged 30-day risk precision from $92.0\%$ to $97.5\%$, outperforming the baseline in every weekly cohort
($p<10^{-3}$, paired sign test). Text serialization instead reduces precision
to $83.0\%$: without sequence-capability injection, serializing up to
$1{,}000$ events into tens of thousands of low-density tokens dilutes attention
and interferes with the backbone's text reasoning rather than aiding it.
Thus merely appending behavior does not explain SeqLLM's gain. The serialized variant was not deployed.

After launch, SeqLLM and the baseline were deployed concurrently in
production. In the deployed 0.6B--8B cascade, the 0.6B scanner retains
$91.24\%$ of confirmed risky merchants at a $0.4\%$ screening ratio over a
six-day monitoring window (Table~\ref{tab:appendix-scanner-recall}), so the
8B ranker's precision gain is realized on top of near-complete funnel
coverage rather than at its expense. Over the same concurrent window,
SeqLLM's appeal rate is $\sim2\%$ versus the baseline's $12\%$, and its
exoneration rate is $0\%$ versus $8\%$ (zero observed exonerations). Full
details are provided in Appendix~\ref{sec:appendix-baselines}.

\paragraph{Deployment II: behavior-token embeddings for a downstream fraud detector.}
The production baseline is a discriminative transaction-pair model whose
behavior embeddings are learned end-to-end from fraud labels. We replace
their initialization with SeqLLM's pretrained behavior embeddings, leaving
all other features, model components, and serving unchanged. In a concurrent
three-month A/B test with users randomly assigned to two equal arms (billions
of transactions per day; 14-day label maturation), Precision@Top-$0.01\%/0.1\%$
increases by $26.8/7.6$ pp and Recall@Top-$0.1\%/1\%$ by $12.9/33.1$ pp---the
largest improvement achieved across all historical iterations of this
production model. Intuitively, SeqLLM embeddings complement sparse fraud
labels with text-grounded semantics and sequence patterns, providing stronger
representations for long-tail behaviors.

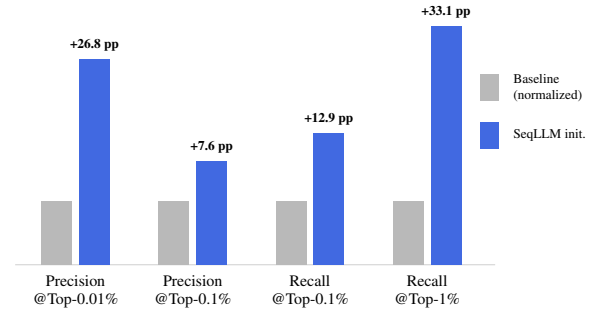
\begin{figure}[htbp]
\centering
\begin{tikzpicture}[x=1cm,y=0.07cm,font=\scriptsize]
  \draw[gray!35] (0,0) -- (6.35,0);

  \fill[gray!55] (0.35,0) rectangle (0.75,12);
  \fill[selfblue] (0.85,0) rectangle (1.25,38.8);
  \node[above,font=\tiny\bfseries] at (1.05,38.8) {+26.8 pp};

  \fill[gray!55] (1.90,0) rectangle (2.30,12);
  \fill[selfblue] (2.40,0) rectangle (2.80,19.6);
  \node[above,font=\tiny\bfseries] at (2.60,19.6) {+7.6 pp};

  \fill[gray!55] (3.45,0) rectangle (3.85,12);
  \fill[selfblue] (3.95,0) rectangle (4.35,24.9);
  \node[above,font=\tiny\bfseries] at (4.15,24.9) {+12.9 pp};

  \fill[gray!55] (5.00,0) rectangle (5.40,12);
  \fill[selfblue] (5.50,0) rectangle (5.90,45.1);
  \node[above,font=\tiny\bfseries] at (5.70,45.1) {+33.1 pp};

  \node[align=center] at (0.80,-5) {Precision\\@Top-0.01\%};
  \node[align=center] at (2.35,-5) {Precision\\@Top-0.1\%};
  \node[align=center] at (3.90,-5) {Recall\\@Top-0.1\%};
  \node[align=center] at (5.45,-5) {Recall\\@Top-1\%};

  \fill[gray!55] (6.15,31) rectangle (6.40,36);
  \node[anchor=west,align=left,font=\tiny] at (6.48,33.5) {Baseline\\(normalized)};
  \fill[selfblue] (6.15,22) rectangle (6.40,27);
  \node[anchor=west,font=\tiny] at (6.48,24.5) {SeqLLM init.};
\end{tikzpicture}
\caption{Online A/B gains from SeqLLM-initialized behavior embeddings. Blue
bars show absolute percentage-point gains over the baseline; absolute values
are withheld for business confidentiality.}
\label{fig:v2-transaction-pair}
\end{figure}

\FloatBarrier
\subsection{Generalization and Comparison with Strong Public Baselines (RQ4)}
\label{sec:v2-generalization}

We assess generalization along three dimensions: MovieLens/Amazon test
transfer from payment behavior to standard recommendation; RecIF provides a
stage-matched comparison with OpenOneRec; and our constructed RecProbe probes
semantic understanding and preference reasoning beyond ranking accuracy.

\begin{table*}[htbp]
\centering
\footnotesize
\caption{Public recommendation and language results. Recommendation
baselines from User-LLM~\cite{ning2025user}; language scores reproduced by us.}
\label{tab:v2-public}
\renewcommand{\arraystretch}{1.0}
\setlength{\tabcolsep}{3pt}
\resizebox{\textwidth}{!}{%
\begin{tabular}{@{}lcccccccc@{}}
\toprule
\multirow{2}{*}{\textbf{Method}}
& \multicolumn{2}{c}{\textbf{MovieLens-20M}}
& \multicolumn{3}{c}{\textbf{Amazon Review}}
& \multicolumn{3}{c}{\textbf{Language Ability}} \\
\cmidrule(lr){2-3} \cmidrule(lr){4-6} \cmidrule(lr){7-9}
& \textbf{Next-item (R@5)}
& \textbf{Fav.\ genre (Acc.)}
& \textbf{Next-item (R@5)}
& \textbf{Fav.\ category (Acc.)}
& \textbf{Review (ROUGE)}
& \textbf{MMLU} & \textbf{C-Eval} & \textbf{AGIEval} \\
\midrule
Vanilla-Sequence
& 0.152 & 0.372 & 0.050 & 0.437 & --
& -- & -- & -- \\
Textualized
& 0.140 & 0.787 & 0.042 & 0.885 & 22.82
& -- & -- & -- \\
User-LLM {\scriptsize (baseline)}
& 0.154 & 0.787 & 0.047 & 0.890 & 26.38
& 0.297 & 0.276 & 0.307 \\
\midrule
\rowcolor{ImproveBlue}
SeqLLM
& \textbf{0.174}\,{\tiny\gc{(+13.0\%)}}
& \textbf{0.973}\,{\tiny\gc{(+23.6\%)}}
& \textbf{0.062}\,{\tiny\gc{(+31.9\%)}}
& \textbf{0.987}\,{\tiny\gc{(+10.9\%)}}
& \textbf{28.62}\,{\tiny\gc{(+8.5\%)}}
& \textbf{0.738} & \textbf{0.786} & \textbf{0.657} \\
\midrule
SeqLLM w/o SemInit
& 0.169 & \underline{0.955} & 0.052 & \underline{0.956}
& \underline{27.99} & 0.737 & 0.786 & 0.656 \\
SeqLLM w/o projector
& \underline{0.172} & 0.934 & \underline{0.061} & 0.925
& 27.23 & 0.735 & 0.784 & 0.655 \\
\bottomrule
\end{tabular}%
}
\end{table*}

\paragraph{Cross-domain transfer on MovieLens and Amazon.}
User-LLM~\cite{ning2025user} is a representative purpose-built baseline that
conditions the LLM on a separate user encoder via cross-attention with the
backbone finetuned jointly. Table~\ref{tab:v2-public} shows SeqLLM outperforms
User-LLM and its Vanilla-Sequence/Textualized controls on all five
recommendation and preference tasks---Recall@5 gains of $13.0\%$ on MovieLens
and $31.9\%$ on Amazon---while retaining substantially higher MMLU/C-Eval/AGIEval.
The two ablations show complementary benefits: semantic initialization helps
next-item prediction, while the projector helps semantic preference and
review generation.

\paragraph{Stage-matched comparison with OpenOneRec on RecIF}
Open OneRec ~\cite{zhou2025openonerec} grounds itemic tokens and acquires
sequence modeling through continual pretraining. We match its 156K behavioral
sequences and 13M-caption alignment pool. Despite using only 100K
general-instruction examples versus OneRec's $\sim$28.6M general-domain and
general-SFT examples, SeqLLM outperforms OneRec-8B-Pretrain on all six
recommendation and language metrics at the sequence-acquisition stage using
$4.4\times$ fewer GPU-days (Table~\ref{tab:v2-recif}).\footnote{We use the
released OneRec-8B checkpoint; OneReason-8B~\cite{onereason2026} was
unavailable at evaluation time.} For final adaptation, OneRec adds SFT,
general-ability distillation, and RL, whereas SeqLLM adds only RL; SeqLLM+RL
still matches the full OneRec-8B pipeline on P@1 and language ability, improves
P@32 by $14.2\%$, and uses $4.8\times$ fewer GPU-days. See
Appendix~\ref{sec:appendix-recif-data} and \ref{sec:appendix-baselines} for
data construction, training budgets, and GPU-day estimation.

\begin{table}[htbp]
\centering
\footnotesize
\caption{RecIF results: sequential recommendation, language ability, and
efficiency. OneRec-PT/OneRec-full = Pretrain$\to$(SFT$\to$Distill$\to$)RL;
SeqLLM/SeqLLM+RL = Align$\to$SFT($\to$RL). CE = C-Eval; AGI = AGIEval; GPU =
GPU-days. Gray subscripts: relative change vs.\ the OneRec baseline in the same
block (compute-reduction factor for GPU-days).}
\label{tab:v2-recif}
\renewcommand{\arraystretch}{1.05}
\setlength{\tabcolsep}{2.5pt}
\resizebox{\columnwidth}{!}{%
\begin{tabular}{@{}l*{7}{c}@{}}
\toprule
\textbf{Method}
& \textbf{P@1} & \textbf{P@32} & \textbf{R@32}
& \textbf{MMLU} & \textbf{CE} & \textbf{AGI}
& \textbf{GPU} \\
\midrule
OneRec-PT
& 0.0204 & 0.1521 & 0.0235
& 0.7112 & 0.7370 & 0.6080
& 460 \\
\rowcolor{ImproveBlue}
\textbf{SeqLLM}
& \makecell{\textbf{0.0411}\\[-1pt]{\tiny\gc{(+101.5\%)}}}
& \makecell{\textbf{0.2398}\\[-1pt]{\tiny\gc{(+57.7\%)}}}
& \makecell{\textbf{0.0354}\\[-1pt]{\tiny\gc{(+50.6\%)}}}
& \makecell{\textbf{0.7221}\\[-1pt]{\tiny\gc{(+1.5\%)}}}
& \makecell{\textbf{0.7667}\\[-1pt]{\tiny\gc{(+4.0\%)}}}
& \makecell{\textbf{0.6646}\\[-1pt]{\tiny\gc{(+9.3\%)}}}
& \makecell{\textbf{105}\\[-1pt]{\tiny\gc{($4.4\times$ fewer)}}} \\
\midrule
OneRec-full
& \textbf{0.0542} & 0.2101 & 0.0356
& 0.7176 & \textbf{0.7489} & 0.6411
& $\sim$994 \\
\rowcolor{ImproveBlue}
\textbf{SeqLLM+RL}
& \makecell{0.0540\\[-1pt]{\tiny\gc{($-$0.4\%)}}}
& \makecell{\textbf{0.2399}\\[-1pt]{\tiny\gc{(+14.2\%)}}}
& \makecell{\textbf{0.0360}\\[-1pt]{\tiny\gc{(+1.1\%)}}}
& \makecell{\textbf{0.7201}\\[-1pt]{\tiny\gc{(+0.3\%)}}}
& \makecell{0.7485\\[-1pt]{\tiny\gc{($-$0.1\%)}}}
& \makecell{\textbf{0.6465}\\[-1pt]{\tiny\gc{(+0.8\%)}}}
& \makecell{\textbf{206}\\[-1pt]{\tiny\gc{($4.8\times$ fewer)}}} \\
\bottomrule
\end{tabular}%
}
\end{table}

\paragraph{Behavior-token understanding and preference reasoning.}
We construct four RecProbe tasks covering video semantics and user preferences
(Table~\ref{tab:v2-rec_understanding}). SeqLLM substantially outperforms
OneRec-8B on all four---e.g., Video--Topic Matching accuracy rises from $0.465$
to $0.745$ and Video Interest Ranking NDCG@3 from $0.112$ to $0.896$. The three
option-based probes are format-robust, so OneRec-8B's lower accuracies reflect
limited semantic grounding rather than formatting failures; the ranking probe
additionally exposes an instruction-following failure of OneRec-8B
(Appendix~\ref{sec:appendix-recif-data}).  These
results indicate that SeqLLM not only grounds new behavior tokens but also
flexibly applies them across tasks---a capability we attribute to the
behavior projector, which provides a shared semantic interface between the new
vocabulary and the LLM. Our controlled projector ablation on industrial data
supports this view: removing the projector preserves single-token translation
but degrades multi-event reasoning and downstream transfer
(Table~\ref{tab:supp-projector-ablation}), consistent with its role as a
compositional interface rather than per-token storage.

\begin{table}[t]
\centering
\small
\caption{Semantic understanding and preference reasoning on RecIF.
Gray text: relative improvement over OneRec-8B.}
\label{tab:v2-rec_understanding}
\renewcommand{\arraystretch}{1.0}
\setlength{\tabcolsep}{4pt}
\resizebox{\columnwidth}{!}{%
\begin{tabular}{@{}llcc@{}}
\toprule
\textbf{Task} & \textbf{Metric} & \textbf{OneRec-8B} & \textbf{Ours} \\
\midrule
\multicolumn{4}{@{}l}{\emph{Video semantics}} \\
Video--Topic Matching       & Acc    & 0.4653 & \textbf{0.7450}\,{\tiny\gc{(+60.1\%)}} \\
Audience Targeting          & Acc    & 0.4733 & \textbf{0.6160}\,{\tiny\gc{(+30.2\%)}} \\
\midrule
\multicolumn{4}{@{}l}{\emph{User preference}} \\
Video Interest Ranking  & NDCG@3 & 0.1121 & \textbf{0.8961}\,{\tiny\gc{($8.0\times$)}} \\
Interest--Category Consistency & Acc   & 0.4973 & \textbf{0.7420}\,{\tiny\gc{(+49.2\%)}} \\
\bottomrule
\end{tabular}%
}
\end{table}

Together, SeqLLM outperforms encoder-based User-LLM and delivers stronger
overall recommendation and token-understanding performance than OpenOneRec at
comparable language ability and substantially lower cost, demonstrating
generalization across datasets, tasks, and sequence--language paradigms
(\textbf{RQ4}).

\subsection{Ablation Summary}
\label{sec:ablation-summary}

Our ablations isolate each design choice: the training objective
(prefix-guided SFT vs.\ CPT, Table~\ref{tab:v2-main_results}), the alignment
stack as a whole (on/off, Table~\ref{tab:v2-main_results}) and its components
individually---semantic initialization and the projector
(Table~\ref{tab:v2-public}, Table~\ref{tab:supp-projector-ablation})---the
input modality (text/behavior/joint, Table~\ref{tab:v2-real-world}), and
behavior serialization (none/text/native tokens, Table~\ref{tab:v2-online-deployment}).

\FloatBarrier

\section{Conclusion}

We presented SeqLLM, a framework that gives pretrained LLMs native
behavioral-sequence modeling while preserving language ability. Its three
components---a compact field-level vocabulary, a text-grounded projector, and
prefix-guided capability injection---match CPT on sequence modeling while
avoiding catastrophic forgetting (C-Eval $0.78$ vs.\ $0.27$ under CPT) and the
need for a separate recovery stage. Deployed at WeChat Pay, SeqLLM raises risk
precision from $92.0\%$ to $97.5\%$ and yields the largest recall gain across
historical iterations of the online fraud-detection model. On public benchmarks it outperforms User-LLM and
OpenOneRec's full pipeline at substantially lower training cost, establishing
capability injection as a scalable route to unified language and
behavioral-sequence models.

\clearpage
\bibliographystyle{ACM-Reference-Format}
\bibliography{acmart.bib}

\clearpage
\appendix

\section{Reproducibility Overview}
\label{sec:appendix-overview}

This appendix provides everything needed to reproduce every result in the main
text, organized by the reproduction workflow: data acquisition and preprocessing
(Appendix~\ref{sec:appendix-data}), model configuration and training with all
baselines (Appendix~\ref{sec:appendix-model}), and per-number computation from a
trained checkpoint (Appendix~\ref{sec:appendix-evaluate}). To avoid duplication,
we do not restate the method (Section~\ref{sec:methodology}) or the metric
definitions (Section~\ref{sec:v2-setup}), reporting only the configurations,
statistics, and protocols required for reproduction.

\begin{table}[htbp]
\centering
\footnotesize
\caption{Mapping from main results to the relevant parts of this appendix.
Training configurations (Appendix~\ref{sec:hyperparameter}) and evaluation
protocols (Appendix~\ref{sec:appendix-evaluate}) are shared across all rows.}
\label{tab:appendix-mapping}
\renewcommand{\arraystretch}{1.15}
\setlength{\tabcolsep}{4pt}

\resizebox{0.8\columnwidth}{!}{%
\begin{tabular}{@{}lll@{}}
\toprule
\textbf{Result} & \textbf{Dataset} & \textbf{Data / baseline setup} \\
\midrule
Tab.~\ref{tab:v2-main_results}      & WeChat Pay        & \S\ref{sec:appendix-wechat-data}, \S\ref{sec:appendix-cpt}, \S\ref{sec:appendix-baselines} \\
Tab.~\ref{tab:v2-real-world}        & WeChat Pay        & \S\ref{sec:appendix-wechat-data}, \S\ref{sec:appendix-baselines} \\
Tab.~\ref{tab:v2-online-deployment} & WeChat Pay        & \S\ref{sec:appendix-baselines} \\
Tab.~\ref{tab:appendix-deployment-candidates} & WeChat Pay & \S\ref{sec:appendix-baselines} \\
Tab.~\ref{tab:appendix-scanner-recall}    & WeChat Pay & \S\ref{sec:appendix-baselines} \\
Tab.~\ref{tab:v2-public}            & MovieLens/Amazon  & \S\ref{sec:appendix-public-data}, \S\ref{sec:appendix-baselines} \\
Tab.~\ref{tab:v2-recif}             & RecIF             & \S\ref{sec:appendix-recif-data}, \S\ref{sec:appendix-baselines} \\
Tab.~\ref{tab:v2-rec_understanding} & RecIF (RecProbe)  & \S\ref{sec:appendix-recif-data} \\
Fig.~\ref{fig:v2-deltaw}            & WeChat Pay        & \S\ref{sec:appendix-cpt} \\
\bottomrule
\end{tabular}%
}
\end{table}

\section{Data and Preprocessing}
\label{sec:appendix-data}

\subsection{WeChat Pay Dataset}
\label{sec:appendix-wechat-data}

\paragraph{Additional sample-construction details.}
All records are anonymized and stripped of personally identifiable information.
For offline evaluation, all merchants appearing in the risk-SFT corpus are
excluded from the held-out test cohorts, making the split entity-disjoint for both
positive and negative labels. Transactions are sorted chronologically before
sequence truncation and tokenization. Table~\ref{tab:appendix-wechat-stats}
consolidates the corpus and stage-level statistics reported in the main text.

\paragraph{Field-level vocabulary.}
To respect confidentiality, individual field names are not disclosed. The full
schema spans $28$ fields (Table~\ref{tab:appendix-wechat-stats}), but not every
field applies to every event: each event type is encoded with its own applicable
field subset, so an event carries about $9$ field tokens on average. The
behavioral vocabulary ($1{,}533$ entries) is the union of all field--value tokens
across the $28$ fields.

\begin{table}[t]
\caption{WeChat Pay corpus statistics. Mean sequence length is computed
after the $1{,}000$-event cap.}
\label{tab:appendix-wechat-stats}
\centering
\scriptsize
\setlength{\tabcolsep}{2.5pt}
\renewcommand{\arraystretch}{1.08}
\resizebox{\columnwidth}{!}{%
\begin{tabular}{@{}lccc@{}}
\toprule
& Train & Validation & Test \\
\midrule
Time span
  & Preceding multi-month window & -- & Subsequent 30-day window \\
Risk-SFT entities (merchants)
  & $4.06$M & -- & $0.99$M/day $\times$ 30 \\
Unlabeled source merchants
  & \multicolumn{3}{c}{$20$M active merchants} \\
Mean / max behavior events
  & 862 / $1{,}000$ & -- & 845 / $1{,}000$ \\
Maximum input tokens
  & \multicolumn{3}{c}{$10{,}000$} \\
Behavior fields / vocab. size
  & \multicolumn{3}{c}{$28$ / $1{,}533$} \\
\midrule
\multicolumn{4}{@{}l@{}}{\emph{Stage sample counts}} \\
\quad Alignment
  & \multicolumn{3}{c}{$117$k ($50$k + $19$k $\times 3$ + $10$k)} \\
\quad Capability injection
  & \multicolumn{3}{c}{$\sim 20$M seq.{} + $\sim 1.1$M text} \\
\quad Risk SFT
  & $4.06$M & -- & $0.99$M/day $\times$ 30 \\
\bottomrule
\end{tabular}%
}
\end{table}

\subsection{MovieLens-20M and Amazon Reviews}
\label{sec:appendix-public-data}

\paragraph{Data sources and filtering.}
MovieLens-20M is the GroupLens 20M-rating release; for Amazon Reviews we use the
\emph{Movies and TV} 5-core subset. Following User-LLM, we retain the users and
items that meet its interaction-count thresholds, sort each user's
interactions by timestamp, and represent an interaction by its metadata: movie
name, genre, and rating for MovieLens, and product title, category, rating, and
review summary for Amazon. Table~\ref{tab:appendix-userllm-data} reports the
resulting user, item, and interaction counts.

\begin{table}[t]
\centering
\footnotesize
\renewcommand{\arraystretch}{0.92}
\setlength{\tabcolsep}{3pt}

\caption{Statistics after the User-LLM preprocessing protocol. The number of test examples equals the number of users.}
\label{tab:appendix-userllm-data}

\begin{tabular}{lrrrr}
\toprule
Dataset & Users & Items & Train & Test\\
\midrule
MovieLens-20M  & 82,977 & 27,280  & 13,821,405 & 82,977\\
Amazon Reviews & 5,756  & 177,978 &   357,258  & 5,756\\
\bottomrule
\end{tabular}
\end{table}

\paragraph{Preprocessing} A sliding window (length $50$) over each user's chronological history yields
supervised examples, where the events preceding a position form the history and
the event at that position is the label. The most recent interaction of every
user is held out for testing and never seen during training, so the test set has
exactly one example per user ($82{,}977$ for MovieLens, $5{,}756$ for Amazon), and
the \emph{Train} column of Table~\ref{tab:appendix-userllm-data} counts the
windowed next-item pairs. Each distinct movie or product is assigned a dedicated
item token, whose readable description (name and genre for MovieLens, title and
category for Amazon) is used for sequence--language alignment
(Section~\ref{sec:behavioral-projector}). At downstream time, a user history is
the chronological sequence of these item tokens, and the three tasks differ only
in the task prefix and target.

\paragraph{Next-item prediction.}
In the prefix-guided capability injection stage, we split each sliding window of length 50 at a ratio of 35:15. Specifically, the first 35 items are taken as the conditional input to generate the subsequent 15 items for multi-item sequence continuation, instead of merely scoring individual candidate items. This operation is applied to both the training and evaluation sets. The model takes the item-token interaction history as input and is optimized to predict the next item token.

\paragraph{Favorite genre / category.}
The label is the genre (MovieLens) or product category (Amazon) that occurs most
often across the user's history, with ties broken by the most recent occurrence.
The input is the same item-token history and the target is the dominant
genre/category, so the task probes long-range preference rather than the last
interaction. Each user yields one instance per split, scored by accuracy.
\begin{seqbox}{Favorite-genre example (MovieLens)}
\exfield{Instruction:} Based on the user's history, what is their favorite
genre?\par\smallskip
\exfield{Input:} <movie\_924> <genre\_103> <ml\_rating\_6> <movie\_919>
<genre\_210> <ml\_rating\_6> $\cdots$ <movie\_3030> <genre\_131>
<ml\_rating\_5>\par\smallskip
\exfield{Output:} <genre\_171> \quad(Horror$|$Mystery$|$Thriller)
\end{seqbox}

\paragraph{Review generation.}
Defined for Amazon only, as MovieLens has no review text: conditioned on the
item-token history and the target product (given by name and rating in the
instruction), the model generates the user's review, and the held-out review of
the test item is the ROUGE reference; empty reviews are discarded.
\begin{seqbox}{Review-generation example (Amazon)}
\exfield{Instruction:} Please write a review for \emph{The Palace of Versailles}.
Your rating is 3.\par\smallskip
\exfield{Input:} <item\_461> <category\_1> <az\_rating\_4> <item\_2649>
<category\_25> <az\_rating\_4> $\cdots$ <item\_51101> <category\_9>
<az\_rating\_3>\par\smallskip
\exfield{Output:} Well done! Biltz was the original person to put \ldots
\end{seqbox}

\subsection{RecIF}
\label{sec:appendix-recif-data}

\paragraph{Data preparation.}
We obtain the data from two official OpenOneRec Hugging Face repositories: the
RecIF dataset\footnote{\url{https://huggingface.co/datasets/OpenOneRec/OpenOneRec-RecIF}}
and the general-purpose SFT
dataset.\footnote{\url{https://huggingface.co/datasets/OpenOneRec/OpenOneRec-General-SFT}}
Following the official preprocessing scripts,\footnote{\url{https://github.com/Kuaishou-OneRec/OpenOneRec}}
we build our splits while faithfully retaining the original semantic IDs (SIDs),
their associated textual descriptions, and the standard train/test partition.
\emph{Alignment data} pair each SID with its textual caption to teach the
projector to map video tokens into language, contain 13$M$+ item-understanding data.
\emph{Capability-injection data} come from the Video-Rec behavioral sequences
(96M interactions from 160K users): to increase sequence supervision, every
training example with a multi-item output is expanded position-wise---each output
position is used in turn as the target of a separate instance while retaining the
associated user-history context (RL variant is optimized on the original,
unexpanded examples).
\emph{Evaluation data} follow the official held-out test split.
We additionally include 100000 general-language examples to preserve language
ability and 5$M$+ item-understanding examples derived from RecIF captions and
SIDs.

\paragraph{RecProbe: item-understanding probing tasks.}
RecProbe is constructed in this work from the public RecIF item captions and
semantic IDs. It contains four probing tasks that evaluate video-token
understanding beyond next-item prediction. Each
task shares the system prompt below and is scored as reported in
Table~\ref{tab:v2-rec_understanding}: Video--Topic Matching, Audience Targeting, and
Interest--Category Consistency use accuracy, and Video Interest Ranking uses NDCG@3. The prompts
are shown in English for presentation; semantic ID token strings
(\texttt{<|sid\_begin|>} $\ldots$ \texttt{<|sid\_end|>}) are verbatim.

We observe that OneRec-8B follows instructions poorly on this task \emph{only}:
rather than producing the requested ranking, it merely re-emits the candidate
SIDs and loops on them, which directly accounts for its low NDCG@3 here. Its
score on this probe therefore reflects grounding and instruction following
jointly, and we scope our claims accordingly (Section~\ref{sec:v2-generalization}).
On the three option-based probes (Video--Topic Matching, Audience Targeting,
and Interest--Category Consistency), OneRec-8B consistently emits a valid
option under the same answer-extraction protocol applied to both models, so its
lower accuracies on those tasks measure semantic understanding rather
than output formatting.

\begin{seqbox}{RecProbe tasks (shared system prompt and one example each)}
\exfield{System:} You are a video semantic-understanding expert who can infer
video content from video tokens.\par\medskip
\exfield{Video--Topic Matching (acc.):} Given videos A--D
(\texttt{<|sid\_begin|>}$\ldots$\texttt{<|sid\_end|>}), which one is most relevant
to the topic ``pets / cute animals''? Answer with the option only.\par\medskip
\exfield{Audience Targeting (acc.):} Given an ad video, which audience is it best
suited for? \emph{(A)} humorous chat screenshots / intimate-relationship
interactions; \emph{(B)} baking and dessert making; \emph{(C)} cartoon characters
and warm emotion; \emph{(D)} game achievements and phone-screen demos. Answer
with the option only.\par\medskip
\exfield{Video Interest Ranking (NDCG@3):} Given the primary interest ``gaming /
esports'' and recent views, rank candidate videos A--C by recommendation priority
(high to low).\par\medskip
\exfield{Interest--Category Consistency (acc.):} Given a behavior sequence and a candidate
video, does the candidate match the sequence's dominant interest?
\emph{(A)} Matches \quad \emph{(B)} Does not match. Answer with the option only.
\end{seqbox}

\begin{seqbox}{Video Interest Ranking: OneRec-8B vs.\ SeqLLM (same input as above)}
\exfield{OneRec-8B:} <|sid\_begin|><s\_a\_5719><s\_b\_513><s\_c\_7881><|sid\_end|>
<|sid\_begin|><s\_a\_5719><s\_b\_2395><s\_c\_3488><|sid\_end|>
<|sid\_begin|><s\_a\_2776><s\_b\_554><s\_c\_8066><|sid\_end|> $\cdots$
{\normalfont\itshape (re-emits candidate SIDs and keeps looping instead of
ranking)}\par\medskip
\exfield{SeqLLM:} ``Answer: Video A > Video B > Video C''
\end{seqbox}

\section{Model, Training, and Baselines}
\label{sec:appendix-model}

The SeqLLM components are defined in Section~\ref{sec:methodology}; this section
adds only reproduction-level configuration and the exact setup of every baseline
and ablation. All settings share the stage order semantic
initialization~$\rightarrow$ translation alignment~$\rightarrow$ reasoning
alignment~$\rightarrow$ prefix-guided injection~$\rightarrow$ downstream tuning,
with the projector and LLM jointly updated unless a baseline requires otherwise.

\subsection{Training Configurations}
\label{sec:hyperparameter}

Table~\ref{tab:all-training-configurations} reports the stage-wise configurations
for all three settings. Behavior-token initialization follows
Section~\ref{sec:behavioral-projector}; tokens whose readable text cannot be
segmented fall back to random initialization.

\subsection{Controlled CPT vs.\ Prefix-Guided SFT Protocol}
\label{sec:appendix-cpt}

The RQ1 comparison (Table~\ref{tab:v2-main_results}, Fig.~\ref{fig:v2-deltaw})
controls every factor except the training objective. Both runs start from the
identical Qwen3-8B checkpoint, share the same corpora, token budget, and
optimizer schedule, and deliberately omit the projector and alignment
curriculum. The sole difference is the objective: CPT uses full-token
next-token prediction on the packed behavior stream, while prefix-guided SFT
computes loss only on the assistant span under a task prefix. Both CPT controls
receive language replay in both raw-text and chat-format renderings (strictly
more language supervision than prefix-guided SFT). Training uses 50k steps,
96 GPUs, DeepSpeed ZeRO-3, and BF16
(Table~\ref{tab:appendix-cpt-hparams}).

\begin{table}[t]
\centering
\caption{Training configuration for CPT vs.\ prefix-guided SFT.
CPT uses same configuration with stage=\texttt{pt} and full-token loss.}
\label{tab:appendix-cpt-hparams}
\small
\setlength{\tabcolsep}{3pt}
\renewcommand{\arraystretch}{0.95}
\begin{tabular}{@{}p{0.35\columnwidth}p{0.58\columnwidth}@{}}
\toprule
\textbf{Setting} & \textbf{Value} \\
\midrule
Backbone
& Qwen3-8B (full FT) \\

General corpora
& Alpaca-GPT4-ZH + Firefly + identity \\

Packing / cutoff
& true / $10{,}000$ \\

Batch size
& 2*2*8*12 \\

Steps
& $50k$ \\

Peak LR
& $2.0\times10^{-5}$ \\

LR schedule
& cosine, warmup $=500$ steps \\

Precision / parallelism
& BF16 / DeepSpeed ZeRO-3 \\

\midrule
CPT objective
& Next-token LM loss on packed stream \\

Prefix-guided SFT
& Assistant-token loss only \\

\bottomrule
\end{tabular}
\end{table}

\begin{table}[!t]
\centering
\footnotesize
\caption{Mechanistic diagnostics of the two objectives, each measured against the
shared Qwen3-8B base (WeChat Pay). Lower weight change / KL and higher similarity
to base indicate less disruption of the pretrained model.}
\label{tab:appendix-cpt-mech}
\renewcommand{\arraystretch}{1.15}
\setlength{\tabcolsep}{5pt}
\begin{tabular}{@{}lcc@{}}
\toprule
\textbf{Diagnostic (vs.\ base)} & \textbf{CPT} & \textbf{Prefix-guided SFT} \\
\midrule
Backbone mean $\lVert\Delta W\rVert_F$ (rel.) & $0.042$ & $\mathbf{0.022}$ \\
Stable rank of $\Delta W$                      & $46.0$  & $\mathbf{26.6}$ \\
Next-token KL from base                        & $0.84$  & $\mathbf{0.21}$ \\
Last-layer CKA to base                         & $0.91$  & $\mathbf{0.98}$ \\
\midrule
Task-vector cosine $\cos(\text{CPT},\text{Prefix-guided SFT})$ & \multicolumn{2}{c}{$0.14$} \\
Next-transaction HR@10                          & $0.802$ & $0.801$ \\
\bottomrule
\end{tabular}
\end{table}

\begin{table*}[t]
\centering
\footnotesize
\caption{Stage-wise training configurations across WeChat Pay,
MovieLens-20M/Amazon Reviews, and RecIF.
T-Align = Translation Alignment; R-Align = Reasoning Alignment;
PGCI = P-G Capability Injection; RC-SFT = Risk-Control SFT;
TS-SFT = Task-Specific SFT; TIA = Text--Item Alignment; RL = Reinforcement
Learning. Global batch sizes are written as per-device batch $\times$
gradient-accumulation steps $\times$ GPUs per node $\times$ nodes; the RL row
reports per-device batch $\times$ nodes.}
\label{tab:all-training-configurations}
\renewcommand{\arraystretch}{1.12}
\setlength{\tabcolsep}{3.2pt}
\resizebox{\textwidth}{!}{%
\begin{tabular}{@{}llcccccc@{}}
\toprule
\textbf{Setting} & \textbf{Stage}
& \textbf{Trainable modules}
& \textbf{LR}
& \textbf{Global batch}
& \textbf{Max length}
& \textbf{Steps/Epochs}
& \textbf{Weight decay} \\
\midrule
\multirow{4}{*}{WeChat Pay}
& T-Align & \multirow{4}{*}{projector + LLM backbone}
& $1\times10^{-5}$ & 4*8*8*12 & 2048 & 2 epochs & 0.01 \\
& R-Align &
& $1\times10^{-5}$ & 4*8*8*12 & 2048 & 2 epochs & 0.01 \\
& PGCI &
& $2\times10^{-5}$ & 2*2*8*12 & 10000 & 50k steps  & 0.01 \\
& RC-SFT &
& $5\times10^{-6}$ & 2*2*8*12 & 10000 & 2 epochs & 0.01 \\
\midrule
\multirow{4}{*}{\makecell[l]{MovieLens-20M\\\& Amazon Reviews}}
& T-Align & \multirow{4}{*}{projector + LLM backbone}
& $1\times10^{-5}$ & 8*16*8*12 & 1024 & 2 epochs & 0.01 \\
& R-Align &
& $1\times10^{-5}$ & 8*16*8*12 & 1024 & 2 epochs & 0.01 \\
& PGCI &
& $2\times10^{-5}$ & 8*16*8*12 & 1024 & 5k steps & 0.01 \\
& TS-SFT &
& $5\times10^{-6}$ & 8*16*8*12 & 1024 & 2k steps & 0.01 \\
\midrule
\multirow{3}{*}{\makecell[l]{RecIF\\(OpenOneRec)}}
& TIA & \multirow{3}{*}{projector + LLM backbone}
& $1\times10^{-5}$ & 8*16*8*12 & 1024 & 1 epoch & 0.01 \\
& PGCI &
& $2\times10^{-5}$ & 2*8*8*12 & 3072 & 2 epochs & 0.01 \\
& RL &
& $1\times10^{-6}$ & 64*12 & 8192 & 1 epoch & - \\
\bottomrule
\end{tabular}%
}
\end{table*}

\subsection{Baselines}
\label{sec:appendix-baselines}

Unless noted otherwise, every baseline shares SeqLLM's datasets, splits,
tokenization, and metrics (Section~\ref{sec:v2-setup}); only the architecture,
input modality, or training objective changes, so each comparison isolates a
single factor. We group baselines by experiment.

\paragraph{Sequence modeling on WeChat Pay (RQ1, Table~\ref{tab:v2-main_results}).}
\textbf{PANTHER}~\cite{li2026panther} is the sequence-only reference (a behavioral
pretraining transformer that outperforms SASRec/HSTU). We retrain it from scratch
on the WeChat Pay corpus with whole-event IDs, matching Qwen3-8B in parameter
count and Transformer configuration and using the capability-injection optimizer,
schedule, and token budget (Table~\ref{tab:all-training-configurations}); it has
no language head, so its language cells are dashes.
\textbf{Qwen3-8B (Text, zero-shot)}~\cite{yang2025qwen3} serializes each
transaction as text (Section~\ref{sec:problem-formulation}) and predicts the next
one; we map its free-form output back to the field vocabulary
(Section~\ref{sec:behavioral-vocabulary}) before scoring HR@10.
\textbf{Qwen3-8B\,+\,CPT (naive)} is the primary matched control (same corpus,
language replay, token budget, and updates as SeqLLM; only the objective differs).

\paragraph{Offline comparison of deployment candidates
(RQ3, Table~\ref{tab:appendix-deployment-candidates}).}
Both DeepSeek-based candidates are fine-tuned on the same downstream
merchant-risk data with the same SFT--RL recipe, optimization setup, and
risk-scoring method. They differ only in sequence input: the text-only variant
receives the merchant profile and complaint text, while the
textualized-sequence variant additionally receives the transaction sequence
serialized as plain text. P@$r\%$ is the risky fraction among the $r\%$
highest-scored merchants, and latency is normalized to SeqLLM under the same
inference stack. Relative changes in Table~\ref{tab:appendix-deployment-candidates}
use the textualized-sequence DeepSeek variant as the reference for each cutoff.

\begin{table}[t]
\centering
\scriptsize
\caption{Complete offline comparison of merchant-ranking deployment
candidates. The textualized-sequence DeepSeek model was not deployed.}
\label{tab:appendix-deployment-candidates}
\renewcommand{\arraystretch}{1.05}
\setlength{\tabcolsep}{2pt}
\resizebox{\columnwidth}{!}{%
\begin{tabular}{@{}lccccc@{}}
\toprule
\textbf{Model} & \textbf{Max tokens} & \textbf{Latency}
& \textbf{P@0.01\%} & \textbf{P@0.1\%} & \textbf{P@1\%} \\
\midrule
\rowcolor{ImproveBlue}
SeqLLM (Ours) & $10{,}000$ & $1\times$
& \makecell{\textbf{0.970}\\{\tiny (+17.7\%)}}
& \makecell{\textbf{0.792}\\{\tiny (+28.6\%)}}
& \makecell{\textbf{0.326}\\{\tiny (+141.5\%)}} \\
DeepSeek-based (text only, SFT+RL) & $40{,}960$ & $20\times$
& \makecell{0.930\\{\tiny (+12.9\%)}}
& \makecell{0.556\\{\tiny ($-9.7\%$)}}
& \makecell{0.231\\{\tiny (+71.1\%)}} \\
DeepSeek-based (text + serialized sequence, SFT+RL) & $40{,}960$ & $20\times$
& \makecell{0.824\\{\tiny (reference)}}
& \makecell{0.616\\{\tiny (reference)}}
& \makecell{0.135\\{\tiny (reference)}} \\
\bottomrule
\end{tabular}%
}
\end{table}

SeqLLM achieves the best precision at all three ranking cutoffs while requiring
one twentieth of the inference time. The textualized-sequence DeepSeek model
degraded performance relative to the text-only DeepSeek model and was
therefore not deployed.

\paragraph{Production merchant risk screening
(RQ3, Table~\ref{tab:v2-online-deployment}).}
The 0.6B and 8B cascade stages share the identical SeqLLM pipeline, differing
only in backbone size (Qwen3-0.6B vs.\ Qwen3-8B;
Table~\ref{tab:all-training-configurations}). The shadow precision evaluates
the 8B ranking stage alone on a shared upstream candidate pool; the 0.6B
scanner was introduced for deployment and is evaluated separately below.
The three-month shadow evaluation, labeling protocol, and appeal/exoneration
definitions are described in Section~\ref{sec:v2-deployment}; here we note only
that exact candidate, action, and appeal counts and traffic composition are
withheld for business confidentiality. The serialized DeepSeek variant was not
deployed.

\paragraph{Recall-side monitoring of the 0.6B scanner
(Table~\ref{tab:appendix-scanner-recall}).}
After deployment, we monitored the 0.6B scanner over a six-day window, scoring
all $\sim$50M active merchants daily. Daily true positives are confirmed
through the same 30-day maturation pipeline as the shadow labels
(Section~\ref{sec:v2-deployment}).
Table~\ref{tab:appendix-scanner-recall} reports Recall@Top-$N$ at three
screening depths. Retaining only the top $0.4\%$ ($\sim$200K candidates) covers
$91.24\%$ of confirmed risky merchants on average. This recall result evaluates
the deployed scanner separately from the 8B-only shadow precision reported in
Table~\ref{tab:v2-online-deployment}.

\begin{table}[t]
\centering
\footnotesize
\caption{Daily recall of the 0.6B scanner over a six-day window. The scanner
scores $\sim$50M merchants per day; true positives are confirmed within a
30-day maturation window. R@Top-$N$ is the fraction of confirmed risky
merchants captured among the $N$ highest-scored merchants.}
\label{tab:appendix-scanner-recall}
\renewcommand{\arraystretch}{1.05}
\setlength{\tabcolsep}{4pt}
\begin{tabular}{@{}lcccc@{}}
\toprule
\textbf{Day} & \textbf{True positives} & \textbf{R@Top-5w} & \textbf{R@Top-10w} & \textbf{R@Top-20w} \\
 & & {\scriptsize(0.1\%)} & {\scriptsize(0.2\%)} & {\scriptsize(0.4\%)} \\
\midrule
1 & 10{,}150 & 0.7778 & 0.8560 & 0.9126 \\
2 & 10{,}046 & 0.7754 & 0.8562 & 0.9123 \\
3 &  9{,}611 & 0.7756 & 0.8515 & 0.9119 \\
4 &  9{,}475 & 0.7747 & 0.8502 & 0.9127 \\
5 &  9{,}355 & 0.7746 & 0.8517 & 0.9123 \\
6 &  9{,}279 & 0.7725 & 0.8510 & 0.9123 \\
\midrule
\textbf{Avg.} & \textbf{9{,}653} & \textbf{0.7751} & \textbf{0.8528} & \textbf{0.9124} \\
\bottomrule
\end{tabular}
\end{table}

\paragraph{Public recommendation on MovieLens/Amazon
(RQ4, Table~\ref{tab:v2-public}).}
\textbf{User-LLM}~\cite{ning2025user} is the main encoder-based competitor. We
reproduce its default \emph{Full} training strategy---the configuration used for
all experiments reported in the original paper: a Transformer user encoder
produces user embeddings that are projected to the LLM hidden size and injected
via cross-attention at intermediate layers, with the Qwen3-8B backbone finetuned
jointly with the encoder, projection, and cross-attention modules. Its lower
language scores in Table~\ref{tab:v2-public} are consistent with this full
finetuning of the backbone on recommendation data. We use the same backbone,
preprocessing, and splits
(Appendix~\ref{sec:appendix-public-data}) and the TS-SFT budget
(Table~\ref{tab:all-training-configurations}). \textbf{Vanilla-Sequence}
(item-token history only) and \textbf{Textualized} (item history as text) reuse
the sequence-only and text-serialization references above.

\paragraph{Item recommendation and understanding on RecIF
(RQ4, Tables~\ref{tab:v2-recif} and~\ref{tab:v2-rec_understanding}).}
\textbf{OneRec-8B}~\cite{zhou2025openonerec} is OpenOneRec's unified
item-token sequence--language model; we use the checkpoint from its official
Hugging Face repository and the official RecIF split
(Appendix~\ref{sec:appendix-recif-data}). At the sequence-acquisition stage, we
compare OneRec-Pretrain with SeqLLM Align$\to$SFT; for final adaptation, we
compare OneRec Pretrain$\to$SFT$\to$Distill$\to$RL with SeqLLM
Align$\to$SFT$\to$RL. Both methods use 156K behavioral sequences and the same
13M-caption alignment pool. OneRec's 33B-token co-pretraining allocates 62\% to
general-domain text ($\sim$26M examples) and is followed by $\sim$2.6M general
SFT examples; SeqLLM uses 100K general-instruction examples.

The GPU-days in Table~\ref{tab:v2-recif} are estimated as
\begin{equation}
\label{eq:gpu-days}
\text{GPU-days}\approx \frac{6ND}{\text{MFU}\cdot F_{\text{peak}}\cdot 86400},
\end{equation}
where $N\!\approx\!8\times10^{9}$, $D$ is the processed-token count (for
OneRec-8B, taken from its officially reported per-stage token
budgets~\cite{zhou2025openonerec}; for SeqLLM, measured from our training
logs), $F_{\text{peak}}$ is the per-GPU BF16 throughput, and
$\text{MFU}\in[0.4,0.5]$. Since both models share
the backbone size and hardware, GPU-days scale with $D$, giving
$460/\!\sim\!994$ (OneRec) vs.\ $105/206$ (SeqLLM), a $4.4\times/4.8\times$
reduction.

\section{Details of Evaluation}
\label{sec:appendix-evaluate}

\paragraph{Next-transaction HR@$k$ (WeChat Pay).} A transaction event is
described by $28$ fields, but next-event prediction is scored on five core
fields that identify a transaction. For each test position we generate the $k$
most probable next events by beam search over the core-field tokens, ranked by
joint probability, and record a hit if the ground-truth event matches one of
the top-$k$ candidates on all five core fields. HR@$k$ is the fraction of
positions with a hit; we report HR@10.

\paragraph{Next-item Recall@$k$ (MovieLens/Amazon).} Each user contributes a
single held-out last item (Appendix~\ref{sec:appendix-public-data}). We rank items
by predicted probability and set Recall@$k=1$ when the held-out item is among the
top-$k$; we report Recall@5. With one relevant item this coincides with HR@$k$.

\paragraph{Top-ranked precision (merchant risk).} The held-out test window
covers 30 consecutive days; on each day we rank the daily cohort
($\approx\!0.99$M merchants) by the risk score $s$
(Appendix~\ref{sec:appendix-baselines}). Precision@Top-$r\%$ is the fraction of
truly risky merchants among the highest-scored $r\%$ of each daily cohort,
averaged over the 30 days. Even at the most selective cutoff ($r=0.01$), this
aggregates $\approx$99 merchants per day, i.e., $\approx$3{,}000 candidates over
the window. We report $r=1/0.1/0.01$ in Table~\ref{tab:v2-real-world}.

\paragraph{Fraud-detection Precision/Recall@Top-$r\%$.}
We conduct a concurrent three-month A/B test with users randomly assigned to
two equal-sized arms. The evaluation covers billions of transactions per day.
Fraud labels return through the production feedback pipeline after a 14-day
maturation window, and only fully matured transactions are included; exact
traffic counts are withheld for business confidentiality.
Within each online A/B arm, we rank transactions by the fraud detection model's
fraud score and select the highest-scored $r\%$. Precision is the fraudulent
fraction within this set, while recall is the fraction of all labeled
fraudulent transactions retrieved by it. Section~\ref{sec:v2-deployment}
reports the SeqLLM arm's absolute percentage-point gains over the production
baseline: precision at $r=0.01/0.1$ and recall at $r=0.1/1$.

\paragraph{RecIF Pass@$k$ and Recall@$k$.} Following
OpenOneRec~\cite{zhou2025openonerec},\footnote{\url{https://github.com/Kuaishou-OneRec/OpenOneRec/tree/main/benchmarks}}
Pass@$k$ measures whether the ground-truth item appears among $k$ generated
candidates and Recall@$k$ the fraction of all relevant items retrieved; we report
P@1, P@32, and R@32.

\paragraph{Preference and understanding metrics.} Favorite-genre/category
prediction and the RecProbe classification tasks
(Appendix~\ref{sec:appendix-recif-data}) are scored by accuracy over the extracted
option; Video Interest Ranking uses NDCG@3 of the three candidates against the
reference order; review generation uses ROUGE-L ($F_1$) between the generated and
held-out reviews.

\paragraph{General language ability.} MMLU, C-Eval, and AGIEval are evaluated with
the lm-evaluation-harness\footnote{\url{https://github.com/EleutherAI/lm-evaluation-harness/}} on vLLM as log-likelihood completion without a chat
template, matching the Qwen3-8B base protocol; C-Eval uses 5-shot, whereas MMLU
and AGIEval are evaluated 0-shot. We report multiple-choice accuracy.

\section{Controlled Ablation of the Behavior Projector}
\label{sec:supp-projector-ablation}

We isolate whether the projector helps the LLM merely memorize new token
meanings or compose and apply them. The full model is compared with a
\emph{w/o-projector} variant (named consistently with
Table~\ref{tab:v2-public}) that removes $g_\psi$ and tunes the behavior-token
embeddings directly. Both variants use the same backbone, text-based token
initialization, translation-to-reasoning data, optimization schedule, and number
of updates; the projector is the sole controlled difference.

We evaluate two capability levels. \emph{Token translation} measures whether the
model can recover the textual description of an individual behavior token.
\emph{Compositional application} measures multi-event reasoning and downstream
risk decisions, where the model must combine token meanings across fields and
time. The w/o-projector variant learns the translation task but transfers
poorly to compositional application. Adding the shared residual projector leaves
token translation intact while substantially improving multi-event reasoning and
downstream transfer. This separation supports the projector's intended role as a
shared semantic interface rather than additional per-token storage.

\begin{table}[t]
\caption{Controlled ablation of the behavior projector on industrial data.
All factors except the projection interface are held fixed. Field Acc.\ is the
fraction of behavior fields translated correctly; higher is better for all
metrics.}
\label{tab:supp-projector-ablation}
\small
\setlength{\tabcolsep}{3.5pt}
\renewcommand{\arraystretch}{1.08}
\begin{tabular}{@{}llcc@{}}
\toprule
\textbf{Capability} & \textbf{Metric}
  & \textbf{w/o projector} & \textbf{w/ projector} \\
\midrule
Token translation
  & Field Acc. & 0.87 & \textbf{0.93} \\
Multi-event Q\&A
  & Accuracy & 0.32 & \textbf{0.86} \\
Risk-assessment Q\&A
  & Accuracy & 0.21 & \textbf{0.78} \\
\bottomrule
\end{tabular}
\end{table}

\end{document}